\documentclass[11pt]{article}

\usepackage[preprint]{acl}

\usepackage{times}
\usepackage{latexsym}

\usepackage[T1]{fontenc}

\usepackage[utf8]{inputenc}

\usepackage{microtype}

\usepackage{inconsolata}

\usepackage{graphicx}

\usepackage{csvsimple}
\usepackage{caption}
\usepackage{subcaption}
\usepackage{multirow}
\usepackage{booktabs}
\usepackage{float}
\usepackage{booktabs}
\usepackage{ulem}

\title{SindBERT, the Sailor: Charting the Seas of Turkish NLP}

\author{
 \textbf{Raphael Schmitt\textsuperscript{1,2}} and
 \textbf{Stefan Schweter\textsuperscript{3}}
\\
 \textsuperscript{1}School of Computation, Information and Technology, Technical University of Munich, Germany,\\
 \textsuperscript{2}Institute of General Practice, Faculty of Medicine and Medical Center, University of Freiburg, Germany, \\
 \textsuperscript{3}Independent Researcher, Holzkirchen, Germany
\\
 \small{
   \textbf{Correspondence:} \href{mailto:raphael.schmitt@uniklinik-freiburg.de}{raphael.schmitt@uniklinik-freiburg.de}
 }
}

\begin{document}
\maketitle
\begin{abstract}
Transformer models have revolutionized NLP, yet many morphologically rich languages remain underrepresented in large-scale pre-training efforts.
With SindBERT, we set out to chart the seas of Turkish NLP, providing the first large-scale RoBERTa-based encoder for Turkish. 
Trained from scratch on 312~GB of Turkish text (mC4, OSCAR23, Wikipedia), SindBERT is released in both base and large configurations, representing the first large-scale encoder-only language model available for Turkish.  
We evaluate SindBERT on part-of-speech tagging, named entity recognition, offensive language detection, and the \textsc{TurBLiMP} linguistic acceptability benchmark.  
Our results show that SindBERT performs competitively with existing Turkish and multilingual models, with the large variant achieving the best scores in two of four tasks but showing no consistent scaling advantage overall.
This flat scaling trend, also observed for XLM-R and EuroBERT, suggests that current Turkish benchmarks may already be saturated.
At the same time, comparisons with smaller but more curated models such as BERTurk highlight that corpus quality and diversity can outweigh sheer data volume.
Taken together, SindBERT contributes both as an openly released resource for Turkish NLP and as an empirical case study on the limits of scaling and the central role of corpus composition in morphologically rich languages.
The SindBERT models are released under the MIT license and made available in both fairseq and Huggingface formats.
\end{abstract}

\section{Introduction}\label{sec:introduction}
The advent of transformer-based models such as BERT~\citep{devlin_bert_2019} and RoBERTa~\citep{liu_roberta_2019} has reshaped natural language processing (NLP), providing contextualized word representations that generalize across a wide range of tasks. While early efforts focused on English and multilingual approaches, research has consistently shown that monolingual pre-training on large, high-quality corpora yields superior results for the target language~\citep{delobelle_robbert_2020,scheible-etal-2024-gottbert,scheible-schmitt-frei-2025-geistbert}.  


For Turkish NLP, several transformer-based encoders have been introduced in recent years. Notable examples include BERTurk~\citep{schweter_berturk_2025}, trained on a 35 GB corpus of Turkish OSCAR, Wikipedia, and OPUS data; ELECTRA~\cite{clark2020electrapretrainingtextencoders} and ConvBERT~\cite{jiang2021convbertimprovingbertspanbased} models trained on both OSCAR and mC4 (35–242 GB)~\cite{jiao2020tinybertdistillingbertnatural}. While these models provide important milestones, most are relatively small encoder models trained with earlier-generation methods or focus on architectures other than RoBERTa. The only RoBERTa models out there were not computed in its fullest extend, but rather with small batch size for relatively small period~\cite{Toraman_2023, tas2024roberturkadjustingrobertaturkish}. Futher, Turkish still lacks a large-scale, high-quality encoder-only model.

To address this gap, we introduce SindBERT, a RoBERTa-based encoder model pre-trained specifically for Turkish. SindBERT builds on the design principles of the German model GottBERT~\citep{scheible-etal-2024-gottbert} and adapts them to the morphological richness and agglutinative structure of Turkish. We construct a byte-level BPE vocabulary optimized for Turkish, train both base and large variants with fairseq~\citep{ott_fairseq_2019}, and leverage TPUv4 hardware~\citep{jouppi_tpu_2023} for efficient large-scale pre-training. SindBERT is designed to combine scalability and reproducibility while directly targeting Turkish, resulting in the first large-scale RoBERTa-style encoder model for Turkish.
Our contributions are as follows:
\begin{itemize}
\item We release SindBERT\textsubscript{base} and SindBERT\textsubscript{large}, trained from scratch on Turkish web-text.
\item We benchmark SindBERT against existing Turkish and multilingual models.
\end{itemize}

Overall, SindBERT is a large-scale Turkish RoBERTa encoder trained from scratch and released openly, showing competitive performance across standard Turkish NLP benchmarks and serving as a robust resource for research and applications; the models are publicly available under the MIT License\footnote{\url{https://huggingface.co/SindBERT}}.



\section{Related Work}
The introduction of transformer-based language models such as BERT~\citep{devlin_bert_2019} and RoBERTa~\citep{liu_roberta_2019} marked a paradigm shift in NLP, enabling significant improvements across a wide range of tasks. Building on these foundations, multilingual extensions such as mBERT and in particular XLM-RoBERTa~\citep{chan_xlm-roberta_2020} became widely used as strong general-purpose baselines across more than 100 languages. At the same time, a wave of monolingual adaptations demonstrated that language-specific pre-training often outperforms multilingual alternatives when sufficient high-quality data is available~\citep{delobelle2020robbert,martin_camembert_2020,chan-etal-2020-germans,scheible-etal-2024-gottbert,scheible-schmitt-frei-2025-geistbert}.  

Recently, multilingual encoder-only models have seen a revival. EuroBERT~\citep{boizard2025eurobertscalingmultilingualencoders} revisits the encoder paradigm with innovations from decoder-only models, introducing a family of multilingual encoders for European and global languages with native support for sequences up to 8,192 tokens. Similarly, mmBERT~\citep{marone2025mmbertmodernmultilingualencoder} scales encoder pretraining to 3T tokens across 1,800+ languages, introducing novel sampling schedules and showing strong performance on both high- and low-resource languages. These developments highlight that encoder-based architectures remain competitive even in an era dominated by large decoder models.

For Turkish, the first widely adopted transformer encoder was BERTurk~\citep{stefan_schweter_2020_3770924}, trained on a 35~GB mixture of OSCAR, Wikipedia, OPUS, and additional resources. Variants included cased/uncased models and vocabularies of 32k or 128k tokens. Distilled versions (DistilBERTurk)~\citep{jiao2020tinybertdistillingbertnatural} and subsequent models such as ELECTRA~\citep{clark2020electrapretrainingtextencoders} and ConvBERTurk expanded the model zoo, with some trained on the Turkish portion of mC4 (up to 242~GB)~\cite{schweter_berturk_2025}. These provided important baselines but generally followed smaller encoder configurations or explored alternative pre-training architectures rather than scaling RoBERTa. 

Building on this line of work, RoBERTurk~\citep{tas2024roberturkadjustingrobertaturkish} introduced a RoBERTa-style encoder specifically adapted for Turkish, showing that refined pre-training objectives and tokenizer design can yield competitive results.
In parallel, research has underscored the critical role of tokenization in morphologically rich languages. \citet{Toraman_2023} systematically analyzed the impact of vocabulary size and segmentation strategy, showing that larger vocabularies can notably improve performance in morphosyntactic evaluations.
However, all these RoBERTa-based models were not extensively trained, typically using moderate batch sizes and relatively few update steps, resulting in comparatively shallow pretraining regimes.


Taken together, these contributions highlight steady progress in Turkish NLP. However, despite the availability of increasingly large corpora and modern training infrastructure, Turkish has lacked a RoBERTa-based encoder model trained from scratch at scale. SindBERT addresses this gap by providing the first large-scale RoBERTa encoder dedicated to Turkish, trained on modern corpora and released openly to the community.

An overview of existing Turkish transformer-based language models is provided in Table~\ref{tab:turkish_models}.


\begin{table*}[htb]
\centering
\begin{tabular}{lp{3cm}p{6.5cm}r}
\hline
\textbf{Model} & \textbf{Architecture} & \textbf{Pre-training Data} & \textbf{Corpus Size} \\
\hline
BERTurk\textsubscript{32k,128k} & BERT base       & OSCAR, Wikipedia, OPUS, non-public & 35 GB \\
DistilBERTurk                   & DistilBERT      & Distilled from BERTurk (subset) & 7 GB \\
ELECTRA\textsubscript{small}    & ELECTRA small   & OSCAR, Wikipedia, OPUS, non-public          & 35 GB \\
ELECTRA\textsubscript{base}     & ELECTRA base    & OSCAR, Wikipedia, OPUS, non-public          & 35 GB \\
ELECTRA\textsubscript{mC4}      & ELECTRA base    & mC4                             & 242 GB \\
ConvBERTurk                     & ConvBERT base   & OSCAR, Wikipedia, OPUS, non-public          & 35 GB \\
ConvBERTurk\textsubscript{mC4}  & ConvBERT base   & mC4                             & 242 GB \\
RoBERTurk & RoBERTa-mid (12L, 1024H)  & OSCAR, Turkish C4 subset (1 GB) & 28 GB \\
SindBERT\textsubscript{base}    & RoBERTa base    & mC4, OSCAR23, Wikipedia         & 312 GB \\
SindBERT\textsubscript{large}   & RoBERTa large   & mC4, OSCAR23, Wikipedia         & 312 GB \\

\hline
\end{tabular}
\caption{Overview of models evaluated in this work. We only consider \textbf{cased} variants even if uncased versions exist.}
\label{tab:turkish_models}
\end{table*}

\section{Methods}
\label{sec:methods}
\subsection{Training Data}
SindBERT was trained on three Turkish corpora: Wikipedia, OSCAR23~\cite{jansen2022perplexedqualityperplexitybasedmethod}, and mC4.  
The corpus was shuffled and lightly filtered, restricted to the removal of documents containing invalid character encodings.
The extracted sizes are approximately 242~GB for mC4, 69~GB for OSCAR, and 0.6~GB for Wikipedia, resulting in a combined pre-training corpus of about 312~GB of Turkish text.

\subsection{Pre-processing}
Similar to RoBERTa, SindBERT relies on byte pair encoding (BPE)~\cite{radford_language_2019} for subword segmentation, which directly operates on raw text without the need for pre-tokenization or auxiliary tools such as Moses~\cite{koehn_moses_2007}. Since the original GPT-2 tokenizer was designed for English, we instead constructed a tokenizer tailored for Turkish. Following the strategy applied in GottBERT~\cite{scheible-etal-2024-gottbert}, we trained a dedicated vocabulary using 40 GB of randomly sampled Turkish text, resulting in a 52k subword inventory optimized for the language. In our experience, sampling around 40 GB of text is already enough for the subword statistics to stabilize, while scaling vocabulary training to the entire corpus would primarily increase computational cost without offering substantial gains. While we did not separately evaluate the effect of this adaptation on storage size or downstream accuracy, previous work in Dutch~\cite{delobelle_robbert_2020} and German~\cite{scheible-etal-2024-gottbert} indicates that language-specific tokenizers can yield improvements in both efficiency and performance.

\subsection{Pre-training}
Following the setup of GottBERT, we pre-trained both SindBERT\textsubscript{base} and SindBERT\textsubscript{large} using the fairseq framework on a 128-core TPUv4 pod~\cite{jouppi_tpu_2023}. Mixed-precision training (fp16/bfloat16) was not employed, so both models were trained entirely in full precision (fp32). This ensures that training dynamics can be attributed directly to model size, without numerical precision optimizations acting as additional factors.


SindBERT\textsubscript{base} completed training in approximately 29.2 hours, while SindBERT\textsubscript{large} required around 6.0 days. We followed the standard RoBERTa pretraining schedule with 100k update steps, a global batch size of 8k, a 10k-step warmup, and polynomial learning rate decay. The base model used a peak learning rate of 0.0004, and the large model 0.00015. Similar to GottBERT~\cite{scheible-etal-2024-gottbert}, we evaluated after each epoch and stored checkpoints throughout training. Since the dataset size only permitted roughly four epochs, the final checkpoint coincided with the best-performing one.

\subsection{Downstream Tasks}
To assess the capabilities of SindBERT, we fine-tuned the model on a diverse suite of Turkish downstream benchmarks covering sequence labeling, text classification, and linguistic acceptability.
Training was performed with the Flair framework~\citep{akbik_flair_2019} v0.15.1, using standardized experiment configurations provided in the repository.  
Hyperparameter optimization was carried out over batch size and learning rate (Table~\ref{tab:hyperparams}), with training capped at a maximum of 30 epochs and early stopping applied (patience = 3).  
All models employed a linear learning rate schedule with a 10\% warmup phase.  
We evaluated SindBERT on the following tasks:

\paragraph{Part-of-Speech Tagging}
We used the concatenation of five Turkish Universal Dependencies (UD)~\cite{nivre-etal-2020-universal} datasets: Atis\footnote{\url{https://github.com/UniversalDependencies/UD_Turkish-Atis}}, BOUN~\cite{ozates-etal-2024-dependency}, FrameNet~\cite{marsan-etal-2021-building}, IMST~\cite{sulubacak-etal-2016-universal}, and Tourism~\footnote{\url{https://github.com/UniversalDependencies/UD_Turkish-Tourism}}.  
This diverse set reflects different domains such as spoken language, newswire, and tourism.  
Providing a measure of syntactic and morphological coverage, we report model's performance using micro F1.


\paragraph{Named Entity Recognition}
For NER, we fine-tuned on the Turkish NER dataset introduced in the WikiANN corpus~\citep{pan-etal-2017-cross} and widely used for multilingual evaluation. We used the splits from \citet{rahimi-etal-2019-massively} and report micro F1 across all entity types.

\paragraph{Offensive Language Detection}
To evaluate robustness on user-generated content, we employed the OffensEval-TR 2020 dataset~\citep{coltekin2020lrec}, a corpus of Turkish tweets annotated for the presence of offensive language. The dataset contains over 31k training and 3.5k test instances, labeled in a binary fashion as either \textit{NOT} (not offensive) or \textit{OFF} (offensive). Mentions and URLs were anonymized during preprocessing (e.g., replaced by @USER or URL), while the tweets otherwise preserve the linguistic and pragmatic properties of social media text.
We report performance using macro F1.

\paragraph{Linguistic Acceptability}
To assess fine-grained grammatical knowledge, we include evaluation on \textsc{TurBLiMP}~\citep{başar2025turblimpturkishbenchmarklinguistic}, a benchmark of 16 core linguistic phenomena ranging from anaphor agreement and argument structure to scrambling and suspended affixation.
Each phenomenon is represented by 1{,}000 minimal pairs, and models are scored following the BLiMP protocol~\citep{warstadt-etal-2020-blimp-benchmark}, i.e., assigning higher probability to the grammatical sentence of each pair.  
For each model we compute the accuracy within every phenomenon and report the average across all 16 categories as the overall \textsc{TurBLiMP} score.  
This measure complements PoS tagging, NER, and sentiment classification by probing deeper syntactic and morphosyntactic competence.

\subsection{Hyperparameters}
We focused our grid search on batch sizes and learning rates, selected based on the most frequent best-performing values in prior experiments (GottBERT, GeistBERT~\cite{scheible-schmitt-frei-2025-geistbert}; see Table~\ref{tab:hyperparams}). Training was applied to PoS, NER and classification and capped at a maximum of 30 epochs, with early stopping applied using a patience of three epochs. All models employed a linear learning rate schedule with a warmup phase of 10\% of the total training steps. All downstream fine-tuning experiments were conducted with a fixed random seed of 1 for the base models and 42 for the large models.
This setup ensures reproducibility and consistency within each scale while maintaining overall comparability across model groups; nonetheless, minor deviations may still arise from seed-related variance~\cite{dodge2020finetuningpretrainedlanguagemodels}.

\begin{table}[h]
\centering
\begin{tabular}{lp{4.5cm}}
\hline
\textbf{Parameter} & \textbf{Values} \\
\hline
Batch Size & 16, 32 \\
Learning Rate & 5e-6, 7e-6, 1e-5, 2e-5, 5e-5 \\
Epochs & up to 30  \newline (Early stopping, patience = 3)  \\
\hline
\end{tabular}
\caption{\label{tab:hyperparams}
Hyperparameter configurations for downstream fine-tuning. Each model–task combination was trained with all permutations, yielding 10 runs per model and task. Reported scores are averaged across seeds for the best configuration.}
\end{table}

\subsection{Model Properties}
Table~\ref{tab:params_tr} summarizes the vocabulary sizes and parameter counts of the Turkish and multilingual models included in our evaluation.  
The smallest encoder is ELECTRA\textsubscript{small} (13.7M parameters), followed by DistilBERTurk (67M).  
Base-scale Turkish encoders, such as ConvBERTurk (cased and mC4 variants), ELECTRA\textsubscript{base} (cased and mC4), and BERTurk (cased/uncased), cluster between 106M and 111M parameters with 32k vocabularies.
RoBERTurk, another RoBERTa-style encoder with a 50k vocabulary, is slightly larger at 125M parameters.
SindBERT\textsubscript{base} grows further to 126M owing to its 52k vocabulary and extended RoBERTa design.

At the mid-scale, mBERT has 178M parameters with a WordPiece vocabulary of nearly 120k tokens, while the 128k-token BERTurk variants reach 184M.  
Among larger models, XLM-R\textsubscript{base} contains 278M parameters, while SindBERT\textsubscript{large} grows to 357M.  
The largest encoder considered is XLM-R\textsubscript{large}, with 560M parameters and a 250k-token vocabulary.  
All values were extracted using Hugging Face’s \texttt{transformers} library.


\begin{table}[!htbp]
\caption{Vocabulary size and total parameter count for Turkish transformer-based models.  
Values were extracted using Hugging Face’s \texttt{transformers} library.}\label{tab:params_tr}
\centering\small
\begin{tabular}{lrr}%
    \hline
    \bfseries Model & \bfseries Vocab Size & \bfseries \#Params \\
    \hline
    \csvreader[late after line = \\]{params.csv}{}%
     {\csvcoli & \csvcolii & \csvcoliii}
    \hline
\end{tabular}
\end{table}

\section{Results}

\subsection{Pre-training}

\begin{figure}[h!tb]
    \centering
    \begin{subfigure}[b]{0.5\textwidth}
         \centering
        \includegraphics[width=\columnwidth]{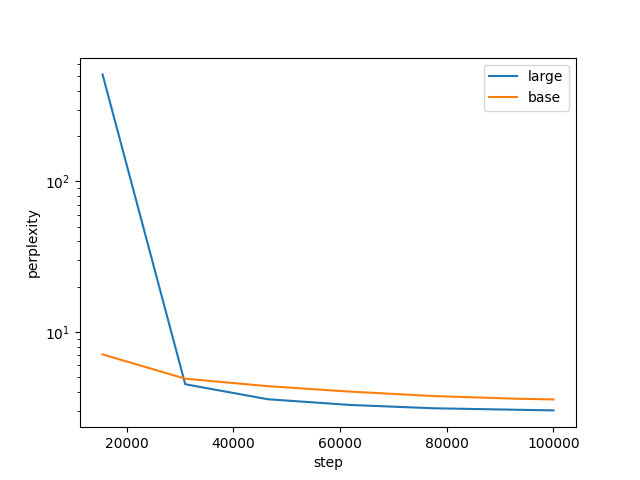}
    \end{subfigure}
    \begin{subfigure}[b]{0.5\textwidth}
         \centering
         \includegraphics[width=\columnwidth]{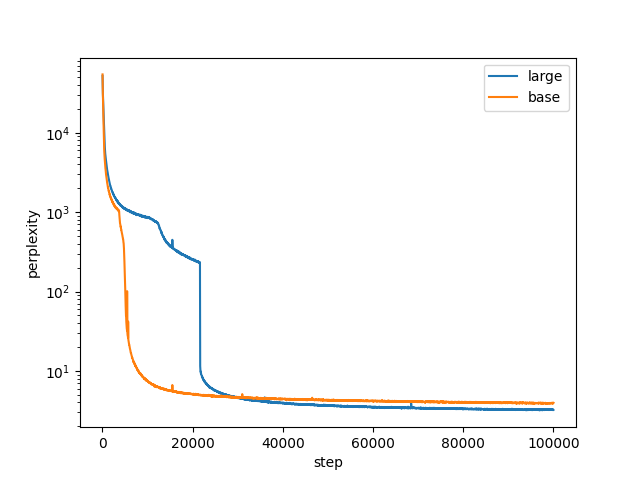}
    \end{subfigure}
    \caption{\label{fig:perplexity}Perplexity of the SindBERT models. Top: validation perplexity measured at checkpoints. Bottom: training perplexity measured at each optimization step.}
\end{figure}

During pre-training, we monitored perplexity both on the training set (at each optimization step) and on the validation set (after each epoch; see Figure~\ref{fig:perplexity}). Across all configurations, the curves follow a consistent convergence pattern. An initial plateau phase can be observed, which is relatively brief for the base models but more pronounced for the large ones. Occasional short upward spikes appear in the training curves; if taken in isolation, these might be misread as divergence, yet they quickly subside as training progresses. 

The base models typically stabilize after 20k–30k steps, while the large models require slightly longer but consistently converge by around 40k steps. By the end of training, both configurations achieve comparably low perplexity, underscoring the efficiency of the pre-training setup. This trend is mirrored in the validation perplexity, which shows steady improvements after each epoch. Overall, training perplexity decreased from about 54.5k to 3.93 for the base models and from about 52.2k to 3.24 for the large models, reflecting robust and reliable convergence.

\subsection{Downstream Tasks}

\paragraph{Part-of-Speech Tagging}

Across base-scale models, performance on the Turkish Universal Dependencies treebank is consistently high, with micro-F1 values exceeding 93\% for nearly all encoders.
The strongest overall results are achieved by ConvBERTurk\textsubscript{mC4} (94.57), closely followed by SindBERT\textsubscript{base} (94.47) and BERTurk\textsubscript{128k} (94.44).
Interestingly, both ConvBERTurk variants, trained with different corpora, maintain a narrow margin over ELECTRA-based and RoBERTa-style encoders, suggesting that architectural innovations like dynamic convolution offer slight but consistent gains in token-level syntactic tagging.
The relatively low score of RoBERTurk (87.99) indicates the limitations of early RoBERTa replications for Turkish, likely due to smaller corpora and shorter training schedules.
SindBERT\textsubscript{base} performs competitively within this saturated range, demonstrating strong generalization across tasks despite a larger 52k BPE vocabulary.

Among large-scale encoders, SindBERT\textsubscript{large} attains the highest F1 (94.63), marginally outperforming XLM-R\textsubscript{large} (94.39).
This indicates that SindBERT’s pre-training on modern Turkish data contributes positively to syntactic coverage, even when compared to substantially larger multilingual models.
The weaker performance of EuroBERT\textsubscript{610M} (93.33) may reflect its more domain-diverse, less Turkish-focused corpus composition.

Overall, POS tagging performance appears saturated across both scales, with nearly all base models exceeding 94 F1 and only marginal gains from scaling. SindBERT maintains parity with top-tier baselines, confirming that syntactic coverage in Turkish is largely solved for transformer-based encoders.

\paragraph{Named Entity Recognition}

The best base-scale performance is reached by BERTurk\textsubscript{32k} (94.38), confirming its robustness for token-level classification.
Close behind are ConvBERTurk (94.03) and BERTurk\textsubscript{128k} (93.81), while SindBERT\textsubscript{base} achieves a solid 93.19, comparable to ELECTRA\textsubscript{base} (93.49) and XLM-R\textsubscript{base} (92.9).
This indicates that SindBERT’s RoBERTa-like setup neither clearly surpasses nor lags behind the most established Turkish encoders, suggesting that the NER task may already be approaching an upper limit with current dataset size and annotation quality.

At the large scale, XLM-R\textsubscript{large} slightly leads (94.44), followed closely by SindBERT\textsubscript{large} (93.64).
Given that XLM-R was trained on over 2 TB of multilingual text, this narrow margin underscores the efficiency of SindBERT’s more compact, Turkish-focused pretraining corpus.

In general, NER results reveal minimal separation between base and large encoders, indicating that model size has limited impact once sufficient Turkish data are used. SindBERT performs on par with the strongest monolingual models, underscoring the stability of its representations across token-level semantic tasks.

\paragraph{Offensive Language Detection}

For offensive language classification (OffensEval-TR 2020), we observe more pronounced differences between architectures.
ConvBERTurk reaches the highest macro-F1 among base models (81.99), with ConvBERTurk\textsubscript{mC4} (81.90) and BERTurk\textsubscript{128k} (81.77) performing almost identically.
ELECTRA variants and SindBERT\textsubscript{base} (81.14) cluster slightly below, while distilled and multilingual models trail more clearly.
These results highlight that models trained on monolingual Turkish corpora still offer clear advantages for pragmatic and domain-sensitive tasks.
SindBERT\textsubscript{base} thus performs solidly but not at the very top, suggesting that further pre-training on informal or social-media text could enhance its stylistic robustness.

In the large model group, SindBERT\textsubscript{large} again performs best (82.29), surpassing XLM-R\textsubscript{large} (81.99) and far exceeding EuroBERT\textsubscript{610M} (75.57).
This consistent lead across two of four downstream tasks emphasizes SindBERT’s balanced architecture and effective use of Turkish-specific corpora.

\begin{table*}[h!tbp]
\centering
\begin{tabular}{lccc|c}
\hline
\textbf{Model} & \textbf{PoS} & \textbf{WikiANN} & \textbf{OffensEval-TR 2020} & \textbf{\textsc{TurBLiMP} AVG} \\
\hline
\csvreader[late after line=\\]{all_base.csv}{}%
 {\csvcoli & \csvcolii & \csvcoliii & \csvcoliv & \csvcolvi}
\hline
\csvreader[late after line=\\]{all_large.csv}{}%
 {\csvcoli & \csvcolii & \csvcoliii & \csvcoliv & \csvcolvi}
\hline
\end{tabular}
\caption{\label{tab:downstream_results}
Evaluation results across four Turkish downstream tasks.  
Best results are shown in bold and second-best results are underlined, with rankings reported separately for base and large model groups. For the 13 base models, third-best results are additionally marked with a dotted underline.
\textbf{PoS}: micro-F1 on concatenated UD datasets.  
\textbf{NER}: entity-level F1 on WikiANN Turkish.  
\textbf{Sentiment}: macro-F1 on OffensEval-TR 2020. 
\textbf{TurBLiMP}: average accuracy over 16 linguistic acceptability phenomena.  
Reported scores for PoS, NER and classification are computed on the test set, with the best checkpoint per model–task combination selected based on validation performance. \textsc{TurBLiMP} was evaluated using its predefined configuration.
}
\end{table*}


\paragraph{\textsc{TurBLiMP}}

Table~\ref{tab:turblimp_detail} reports the detailed \textsc{TurBLiMP} results for all base and large models. 
Overall, SindBERT\textsubscript{base} achieves an average score of 90.3, which is comparable to ELECTRA\textsubscript{base} and ELECTRA\textsubscript{mC4} (both 89.9), while trailing behind the strongest baselines BERTurk\textsubscript{32k} (93.8) and BERTurk\textsubscript{128k} (95.1). 
A closer look at the per-phenomenon results shows that SindBERT\textsubscript{base} is particularly strong on \textit{scrambling},
\textit{suspended affixation}, \textit{subject agreement}, and \textit{irregular forms} (all $\geq$98), which are central morphosyntactic phenomena of Turkish.
At the same time, it struggles with \textit{ellipsis} (59.0) and \textit{island effects} (64.0), two categories that remain challenging across most models.  

For the large models, SindBERT\textsubscript{large} reaches an average of 89.8, placing it slightly below EuroBERT\textsubscript{610M} (90.0) and XLM-R\textsubscript{large} (92.7). 
Its strengths mirror the base variant: ceiling-level performance in morphologically rich categories such as \textit{suspended affixation}, \textit{scrambling}, and \textit{irregular forms}. 
However, SindBERT\textsubscript{large} shows a severe weakness in \textit{ellipsis} (27.8), which strongly lowers its overall average.

These findings highlight that monolingual models like SindBERT capture Turkish-specific morphosyntax particularly well, while multilingual models such as XLM-R generalize more effectively to harder syntactic phenomena (e.g., ellipsis and binding). 
This suggests a trade-off between specialization in language-specific structures and broader generalization capacities learned from multilingual corpora.


\begin{table*}[h!tbp]
\centering
\resizebox{\textwidth}{!}{
\setlength{\tabcolsep}{3pt}
\begin{tabular}{lcccccccccccccccc|c}
\hline
\textbf{Model} & Ana. Agr. & Arg. Tr. & Arg. Ditr. & Bind. & Det. & Ellip. & Irr. & Isl. & Nom. & NPI & Pass. & Quant. & RelCl. & Scramb. & Subj. Agr. & Susp. Aff. & \textbf{AVG} \\
\hline
\csvreader[late after line=\\]{turblimp_detail_base.csv}{}%
 {\csvcoli & \csvcolii & \csvcoliii & \csvcoliv & \csvcolv & \csvcolvi & \csvcolvii & \csvcolviii & \csvcolix & \csvcolx & \csvcolxi & \csvcolxii & \csvcolxiii & \csvcolxiv & \csvcolxv & \csvcolxvi & \csvcolxvii & \csvcolxviii}
\hline
\csvreader[late after line=\\]{turblimp_detail_large.csv}{}%
 {\csvcoli & \csvcolii & \csvcoliii & \csvcoliv & \csvcolv & \csvcolvi & \csvcolvii & \csvcolviii & \csvcolix & \csvcolx & \csvcolxi & \csvcolxii & \csvcolxiii & \csvcolxiv & \csvcolxv & \csvcolxvi & \csvcolxvii & \csvcolxviii}
\hline
\end{tabular}
}
\caption{\label{tab:turblimp_detail}
Detailed \textsc{TurBLiMP} evaluation across 16 linguistic acceptability phenomena. 
Best results are shown in bold and second-best results are underlined, with rankings reported separately for base and large model groups. 
For the 13 base models, third-best results are additionally marked with a dotted underline.}
\end{table*}

\section{Discussion}

\subsection{Principal Findings}
Our evaluation shows that SindBERT\textsubscript{base} performs competitively with other widely used Turkish encoders, confirming the robustness of its RoBERTa-style pretraining setup.
At the same time, SindBERT\textsubscript{large} achieves the best overall results in two of four downstream tasks, notably in part-of-speech tagging and offensive language detection, and also performs strongly on several linguistic control tests.
While scaling does not produce uniform gains across all benchmarks, these task-specific improvements suggest that larger contextual capacity primarily benefits pragmatically and syntactically complex settings.
Similar saturation effects are visible for EuroBERT and XLM-R, indicating that many Turkish benchmarks may no longer be sufficiently discriminative to reveal consistent scaling trends.
Nonetheless, diagnostic evaluations such as \textsc{TurBLiMP} underscore SindBERT’s strengths in Turkish-specific grammatical phenomena (e.g., scrambling, suspended affixation, subject agreement), highlighting the model’s linguistic depth beyond aggregate scores.


\subsection{Corpora}
A likely factor explaining the limited scaling gains lies in the training corpus composition.
SindBERT was trained on 312 GB of text—dominated by mC4 (242 GB), which provides broad coverage but is considerably noisier than smaller, curated datasets.
By contrast, BERTurk, trained on only a fraction of that volume but sourced from cleaner collections (OSCAR, Wikipedia, OPUS, and non-public), achieves excellent results, particularly on linguistically sensitive evaluations.
This mirrors trends observed in other monolingual models such as GottBERT, CamemBERT, and GeistBERT, where performance gains stemmed not merely from data size but from an effective balance of quality, domain diversity, and linguistic representativeness.
Our findings therefore reinforce that corpus curation, not scale alone, is decisive for progress in Turkish NLP.

A further dimension concerns vocabulary design.
SindBERT employs a 52k BPE vocabulary that balances coverage and efficiency, whereas BERTurk also released a 128k-token variant, which ranks among the strongest performers in our benchmarks, especially on \textsc{TurBLiMP}.
Recent work by \citet{Toraman_2023} corroborates that vocabulary size has a substantial impact on Turkish models due to the language’s agglutinative morphology.
They report that optimal vocabulary scales differ by tokenization strategy: for BPE or WordPiece, vocabularies around 20\% of model parameters tend to be most effective, while morphological or word-level tokenizers may benefit from substantially larger ratios.
Our results align with this observation: BERTurk\textsubscript{128k} profits from an expanded vocabulary despite its smaller corpus, whereas SindBERT’s 52k vocabulary remains sufficiently expressive to achieve competitive results given its broader but noisier training data.

\subsection{Efficiency}
From an efficiency perspective, our findings highlight a favorable trade-off between scale and performance.
While SindBERT\textsubscript{base} achieves results comparable to its larger counterpart at a fraction of the computational cost, SindBERT\textsubscript{large} still demonstrates measurable advantages on more demanding or pragmatically complex tasks.
This indicates that the large model’s additional capacity is not wasted, but rather contributes selectively where richer contextual representations are required.
Nevertheless, for most real-world scenarios, the base configuration offers an excellent balance between efficiency and accuracy.
Taken together, the flat scaling behavior across multiple Turkish model families suggests that future progress will hinge less on parameter growth and more on corpus quality, tokenization, and task design.

\section{Future Directions}
Future work may extend SindBERT in several directions. 
First, while GeistBERT built on the GottBERT checkpoint through continued pre-training on in-domain data~\cite{scheible-schmitt-frei-2025-geistbert}, and ChristBERT explored the effects of continued pre-training versus training from scratch using both general and domain-specific vocabularies, a similar ablation study has not yet been conducted for Turkish. 
SindBERT provides a natural starting point for replicating these approaches, enabling systematic comparisons of domain adaptation strategies in Turkish.  

Second, recent work on PortBERT~\cite{scheible-schmitt-etal-2025-portbert} suggests that efficiency considerations, both during downstream fine-tuning and at inference time, deserve closer inspection alongside raw performance. Adopting a similar perspective for Turkish NLP could help assess how different models trade off accuracy against computational cost.

Third, our findings indicate that many existing benchmarks are already saturated, as they fail to reveal consistent improvements from larger models.
To overcome this limitation, future evaluations should adopt more comprehensive and discriminative test suites.
In particular, the recently released TrGLUE benchmark\footnote{\url{https://huggingface.co/datasets/turkish-nlp-suite/TrGLUE}} offers a promising step in this direction, providing a diverse collection of tasks. It includes natural language inference, paraphrase detection, sentiment analysis, and question answering, that more closely mirror the breadth of the original GLUE suite.
Incorporating TrGLUE into future experiments would enable a more fine-grained assessment of SindBERT’s generalization capabilities across both syntactic and semantic dimensions.

Fourth, extending evaluation to specialized domains such as biomedical or legal language remains an important frontier for Turkish NLP, where SindBERT could serve as a foundation for targeted domain adaptation, just as GottBERT~\cite{scheible-etal-2024-gottbert} and GeistBERT~\cite{scheible-schmitt-frei-2025-geistbert} did for ChristBERT~\cite{he_word_2025}.

Finally, future pre-training efforts could further improve linguistic coverage by considering document or sentence boundaries during sampling and by employing WWM~\cite{martin_camembert_2020, chan-etal-2020-germans}.

\section{Conclusion}
We introduced SindBERT, the first large-scale RoBERTa encoder trained from scratch on 312 GB of Turkish text.
Across four benchmarks, it performs competitively with existing models, with SindBERT\textsubscript{large} achieving the best results in two tasks.
While scaling brings only selective gains, this mirrors trends in XLM-R and EuroBERT, suggesting that Turkish benchmarks are nearing saturation.
The contrast with BERTurk highlights the decisive role of corpus quality and variance over size.
Together, these findings show that progress in Turkish NLP will depend less on scaling and more on curated data, adaptive tokenization, and challenging evaluation suites.
As the first openly released large-scale RoBERTa model for Turkish, SindBERT establishes a solid foundation for future Turkish NLP.

\section*{Limitations}
This work has several limitations. First, SindBERT was trained on three large-scale Turkish corpora (mC4, OSCAR23, Wikipedia) with only light filtering applied, restricted to the removal of documents containing invalid character encodings. No additional cleaning, quality filtering, or cross-source deduplication was performed. As a result, residual noise, duplicated content, and potential biases are likely to remain in the training data and may influence the learned representations.  

Second, the training data was drawn exclusively from web-based sources, without explicit control for dialectal or register variation (e.g., Ottoman vs.\ Modern Turkish, formal vs.\ colloquial, or regional varieties). This may limit the model’s robustness on underrepresented varieties or in specialized domains such as biomedical or legal text, unless additional domain-adaptive pre-training is performed.  

Third, SindBERT was pre-trained with conservative hyperparameter settings and without extensive exploration of alternative masking strategies (e.g., Whole Word Masking) or longer training schedules. Pre-training was also conducted without mixed precision, which increased computational cost and limited the feasibility of scaling to larger model sizes or more training steps.

Fourth, we did not perform a systematic error analysis of downstream results. Such an analysis could provide insights into systematic weaknesses (e.g., frequent PoS confusions, NER boundary errors, sentiment misclassifications, or \textsc{TurBLiMP} minimal pair failures) and help prioritize future improvements in model design and dataset composition.  

Fifth, baseline reproducibility introduces some uncertainty. ConvBERTurk and ConvBERTurk\textsubscript{mC4} are based on the ELECTRA codebase, but during conversion from the original checkpoints to HuggingFace Transformers the distinction between generator and discriminator is not explicit. While ELECTRA’s conversion script allows specifying this choice, ConvBERTurk appears to default to the discriminator. This may not invalidate comparisons, but it does leave open the possibility of subtle architectural differences and explains the suboptimal performance on \textsc{TurBLiMP}.

Lastly, our evaluation focused on four downstream tasks (PoS tagging, NER, sentiment classification, \textsc{TurBLiMP}). While these cover a diverse range of morphosyntactic, semantic, and syntactic phenomena, they do not capture the full scope of Turkish NLP challenges such as question answering, natural language inference, summarization, or long-context understanding. The generalization of SindBERT to these settings remains to be established.

\section*{Ethical Considerations}
Like all large-scale language models, SindBERT may inherit biases from its training data, which can influence downstream tasks such as classification or decision-making. While no deduplication was applied, the corpus may still contain redundancy and noise, as well as deeper societal or representational biases. Furthermore, training on large web-based corpora raises privacy concerns, as models may inadvertently retain sensitive information. Responsible deployment is especially important in high-stakes domains like legal, medical, or financial NLP.

Despite optimizations for efficiency, pre-training and evaluating transformer models remain computationally demanding, contributing to energy use and carbon emissions. These environmental costs highlight the need for balancing model performance with sustainable development goals.

\section*{Acknowledgments}
The authors gratefully acknowledges the support of Google’s TPU Research Cloud for providing access to Cloud TPUs, which enabled efficient pretraining of SindBERT. The authors also thank Nora Limbourg, the assigned Google Cloud Customer Engineer, for her valuable technical assistance and coordination throughout the project.
Finally, the authors gratefully acknowledge the scientific support and resources of the AI service infrastructure LRZ AI Systems provided by the Leibniz Supercomputing Centre (LRZ) of the Bavarian Academy of Sciences and Humanities (BAdW), funded by Bayerisches Staatsministerium für Wissenschaft und Kunst (StMWK).


\bibliography{custom}

\newpage
\appendix

\section{Runtime}

Table~\ref{tab:best_params} lists the hyperparameters of the best SindBERT models (selected by validation performance) for each benchmark, supporting reproducibility of our results.  
For transparency, Table~\ref{tab:gpu_time} reports the total computation time per task, showing that all Turkish downstream experiments together required roughly 425 GPU hours (about 17.7~days). All base model experiments were run on an NVIDIA RTX 3090, and large model experiments on an NVIDIA H100 GPU.

\textsc{TurBLiMP} is not reported, as the pipeline did not record training time.
Since no hyperparameter search was involved, this omission is minor and corresponds to only a few additional hours.

\begin{table}[H]
    \centering
    \begin{tabular}{lr}
         \hline
         \bfseries Task & \bfseries Computation Time \\
         \hline
         PoS  & 200:21 \\
         WikiANN & 131:02 \\
         OffensEval-TR 2020 & 93:37 \\
         \hline
         Total & 425:01 \\
         \hline
    \end{tabular}
    \caption{Computation time in hours and minutes for the Turkish downstream tasks, summing to about 425 hours and 1 minute (approximately 17.7 days).}
    \label{tab:gpu_time}
\end{table}

\begin{table*}[h!tbp]
\centering
\begin{tabular}{lcccccc}
    \hline
    \multirow{2}{*}{\textbf{Model}} & 
    \multicolumn{2}{c}{\textbf{PoS}} & 
    \multicolumn{2}{c}{\textbf{NER}} & 
    \multicolumn{2}{c}{\textbf{Sentiment}} \\
    \cmidrule(lr){2-3}\cmidrule(lr){4-5}\cmidrule(lr){6-7}
     & BF & LR & BF & LR & BF & LR \\
    \hline
    \csvreader[late after line = \\]{hyperparams_base.csv}{}%
     {\csvcoli & \csvcolii & \csvcoliii & \csvcoliv & \csvcolv & \csvcolvi & \csvcolvii}
    \hline
    \csvreader[late after line = \\]{hyperparams_large.csv}{}%
     {\csvcoli & \csvcolii & \csvcoliii & \csvcoliv & \csvcolv & \csvcolvi & \csvcolvii}
    \hline
\end{tabular}
\caption{\label{tab:best_params}
Hyperparameters of the best-performing downstream task model for each pre-trained model.  
BF denotes the batch size, LR the learning rate.  
}
\end{table*}

\end{document}